%% file: main.tex
\documentclass[letterpaper]{article} 
\usepackage{aaai24}  
\usepackage{times}  
\usepackage{helvet}  
\usepackage{courier}  
\usepackage[hyphens]{url}  
\usepackage{graphicx} 
\usepackage{natbib}  
\usepackage{caption} 
\usepackage{algorithm}
\usepackage{algorithmic}
\usepackage{booktabs}
\usepackage{multirow}
\usepackage{amsmath}
\usepackage{amssymb}
\usepackage{diagbox}
\usepackage{tikz}
\usetikzlibrary{tikzmark}
\usepackage{newfloat}
\usepackage{listings}
\DeclareCaptionStyle{ruled}{labelfont=normalfont,labelsep=colon,strut=off} 
\floatstyle{ruled}
\newfloat{listing}{tb}{lst}{}
\floatname{listing}{Listing}
\usepackage{enumitem}

\title{Empowering Dual-Level Graph Self-Supervised Pretraining with Motif Discovery}
\author {
    Pengwei Yan\textsuperscript{\rm 1,\rm 2},
    Kaisong Song\textsuperscript{\rm 2, \rm 3},
    Zhuoren Jiang\textsuperscript{\rm 1}\thanks{Corresponding authors.},
    Yangyang Kang\textsuperscript{\rm 2}\footnotemark[1],
    Tianqianjin Lin\textsuperscript{\rm 1,\rm 2},
    Changlong Sun\textsuperscript{\rm 2},
    Xiaozhong Liu\textsuperscript{\rm 4}
}
\affiliations {
    \textsuperscript{\rm 1}Department of Information Resources Management, Zhejiang University, Hangzhou, 310058, China\\
    \textsuperscript{\rm 2}Alibaba Group, Hangzhou, 311121, China\\
    \textsuperscript{\rm 3}Northeastern University, Shenyang, 110819, China\\
    \textsuperscript{\rm 4}Computer Science Department, Worcester Polytechnic Institute, Worcester, 01609-2280, MA, USA\\
    \{yanpw, jiangzhuoren, lintqj\}@zju.edu.cn,
    \{kaisong.sks, yangyang.kangyy\}@alibaba-inc.com,
    changlong.scl@taobao.com,
    xliu14@wpi.edu
}

\begin{document}
\maketitle

\input{contents/0abs}

\input{contents/1intro}
\input{contents/4method}

\input{contents/5experiments}
\input{contents/2relatedWork}

\input{contents/6conclusion}

\section*{Acknowledgements}
This work is supported by the National Natural Science Foundation of China (72104212, 72134007, 62106039), the Natural Science Foundation of Zhejiang Province (LY22G030002), the Fundamental Research Funds for the Central Universities, and Alibaba Group through Alibaba Innovative Research Program.

\bibliography{aaai24}
\input{contents/7appendix}

\end{document}

%% file: contents/0abs.tex
\begin{abstract}
While self-supervised graph pretraining techniques have shown promising results in various domains, their application still experiences challenges of limited topology learning, human knowledge dependency, and incompetent multi-level interactions. To address these issues, we propose a novel solution, Dual-level Graph self-supervised Pretraining with Motif discovery (DGPM), which introduces a unique dual-level pretraining structure that orchestrates node-level and subgraph-level pretext tasks. Unlike prior approaches, DGPM autonomously uncovers significant graph motifs through an edge pooling module, aligning learned motif similarities with graph kernel-based similarities. A cross-matching task enables sophisticated node-motif interactions and novel representation learning. Extensive experiments on 15 datasets validate DGPM's effectiveness and generalizability, outperforming state-of-the-art methods in unsupervised representation learning and transfer learning settings. The autonomously discovered motifs demonstrate the potential of DGPM to enhance robustness and interpretability.
\end{abstract}

%% file: contents/1intro.tex
\section{Introduction}

As in the fields of natural language processing and computer vision \cite{devlin2018bert,he2020momentum}, pretraining with self-supervised learning (SSL) holds similar significance for models constructed on graph data, which aims to learn the informative graph representations from unlabeled data \cite{hu2019strategies}. Such approaches can effectively address the labeled data scarcity and improve the model's generalizability in graph domain \cite{xie2022self}. While numerous research endeavors \cite{velivckovic2018deep,you2020graph,hou2022graphmae} have already yielded successful results, applying self-supervised pretraining techniques to graph data still faces the following challenges.
\begin{figure}
    \centering
    \includegraphics[width=8.5cm]{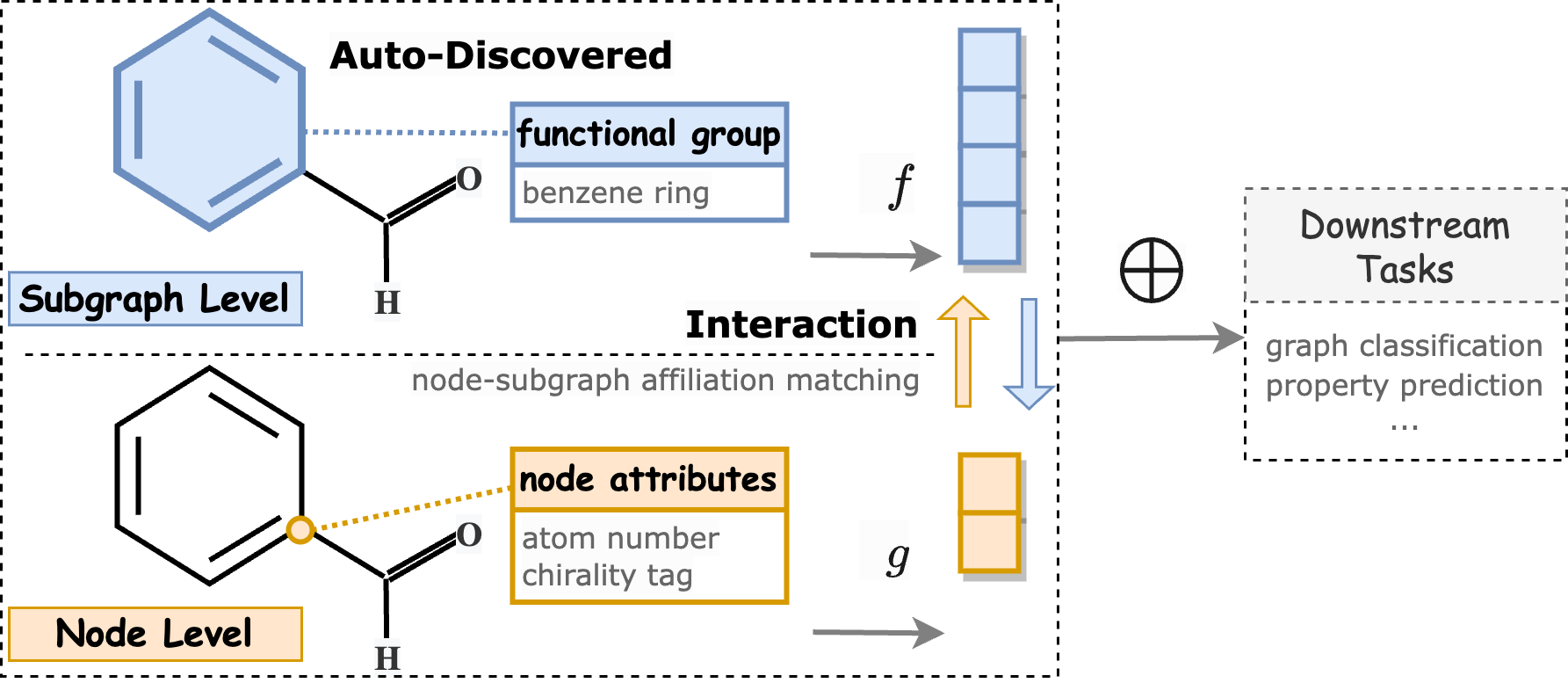}
    \caption{Toy example showing interactive dual-level graph pretraining.}
    \label{fig:toy_example}
\end{figure}

\begin{figure*}
    \centering
    \includegraphics[width=\textwidth]{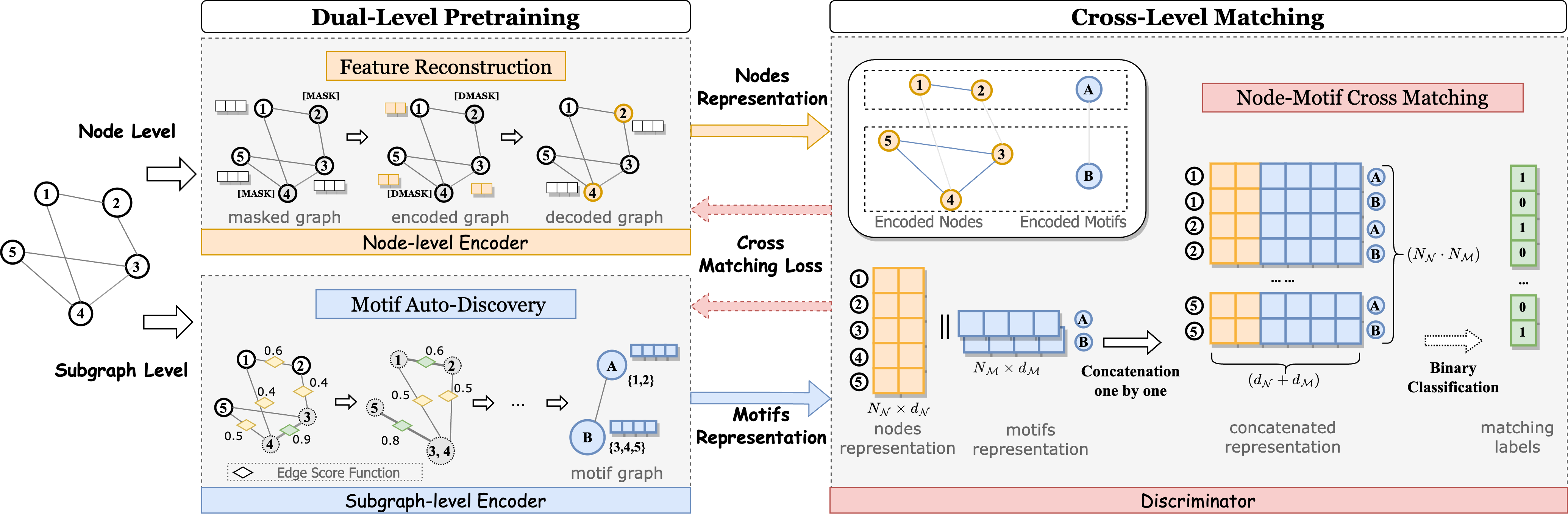}
    \caption{The DGPM framework comprises two main components: Dual-level pretraining and Cross-level matching. The Dual-level pretraining encompasses a node-level feature reconstruction task and a subgraph-level motif auto-discovery task.}
    \label{fig:model_framework}
\end{figure*}


\textbf{Limited Topology Learning}. Defining appropriate pretext training tasks stands as a pivotal objective for graph pretraining models. However, most existing graph pretraining endeavors often restrict themselves to simple prediction tasks rooted in close neighborhood structures. For instance, utilizing k-hop neighborhood information \cite{rong2020self} to predict node contextual property, and employing subgraphs to forecast the neighboring graph structures \cite{zhang2021motif}. However, these learning tasks might be confined to predicting lower-order structural attributes, potentially hampering the model's ability to comprehend graph topology \cite{chen2023motif} and also neglecting higher-order structural interpretability \cite{milo2002network}.

\textbf{Human Knowledge Dependency}. Graph motif \cite{milo2002network}, the significant high-order graph pattern with semantic meaning, have shown their potential to enhance graph models \cite{chen2023motif,rong2020self,zhang2021motif}. Despite their beneficial impact, existing motif-related methods \cite{rong2020self,zhang2021motif} often rely on domain knowledge and manual motif pre-definition, limiting generalizability across domains for motif-driven self-supervised learning.

\textbf{Incompetent Multi-Level Interactions}. For effective pretraining, graph self-supervised learning models should capture knowledge from node attributes and structural topology \cite{ma2021deep}. But existing approaches often use step-by-step learning \cite{hu2019strategies} or simple aggregation learning \cite{zhang2021motif}, missing intrinsic multi-level information interaction.

To address the above challenges, we propose Dual-level Graph self-supervised Pretraining with Motif discovery (\textbf{DGPM}). As shown in Figure~\ref{fig:toy_example}, DGPM introduces a dual-level pretraining architecture. In addition to a node-level pretraining task, DGPM incorporates a novel and challenging task for subgraph-level pretraining. This task involves the autonomous discovery of motifs through an edge pooling module. The training objective is aligning two kinds of similarity: one derived from automatically learned motifs and the other based on graph similarity computed leveraging the Wasserstein Weisfeiler Lehman (WWL) graph kernel. Furthermore, a cross-matching task targeting node-motif affiliations is established to connect the node and subgraph-level pretraining.

Extensive experiments on 15 datasets, including validation of unsupervised representation learning and transfer learning, demonstrate the effectiveness and generalizability of DGPM. In unsupervised representation learning for graph classification, DPGM outperforms all SOTA methods and shows comparable performance with supervised ones. Besides, ablation studies, sensitive analyses, and auto-discovered motif analyses provide further validation of the reliability and robustness of our proposed framework.

The major contributions of this research can be summarized as follows:

1. We propose a novel dual-level graph pretraining architecture, DGPM\footnote{The code is available at https://github.com/RocccYan/DGPM.}, to address the challenges of limited topology learning, human knowledge dependency, and incompetent multi-level interactions for graph self-supervised pretraining. To the best of our knowledge, this is the first graph pretraining framework that utilizes a motif auto-discovery mechanism to leverage subgraph structure information in self-supervised learning.

2. A novel and challenging pretext task for subgraph-level pretraining is introduced to comprehensively learn the vital graph structures. In which, the motif auto-discovery module can autonomously uncover the crucial patterns in graph structure to enhance the generalization of graph pretraining model and improve the interpretability by providing the visualized motifs in the pretraining process.

3. Through extensive experiments on 15 benchmark datasets, we demonstrate the superiority and generalizability of DGPM by comparing it with a number of strong baselines. The learned motif analysis further proves the potential of DGPM to enhance interpretability. To help other scholars reproduce the experiment outcome, we will release the code and datasets publicly.

%% file: contents/4method.tex
\section{Methodology}
Unlike prior research, the proposed DGPM aims to learn both node-level attributes and subgraph-level structures of graphs. The framework of DGPM is illustrated in Figure~\ref{fig:model_framework}.

\subsection{Problem Statement}

\subsubsection{Notations.}
Let $G=(V,E)$ denote a graph,  where ${V} = \{v_1,...,v_{N_{\mathcal{N}}\}}$ represents the set of nodes, and ${E} \subseteq {V \times V}$ represents the set of edges. $G$ is associated with a feature matrix $\mathbf{X} \in \mathbb{R}^{N_{\mathcal{N}}\times d}$, and an adjacency matrix $\mathbf{A} \in \mathbb{R}^{N_{\mathcal{N}}\times N_{\mathcal{N}}}$ where $\mathbf{A}_{ij}=1$ iff $(v_i,v_j) \in \mathcal{E}$ and $\mathbf{A}_{ij}=0$ otherwise.
Following \cite{milo2002network},  
we define motifs $\mathcal{M}=(V_\text{S},E_\text{S})$ as high frequency subgraphs of $G$, with $V_\text{S} \subseteq V$ and $E_\text{S}\subseteq E$. 


\subsubsection{Task: Dual-Level Graph Self-Supervised Pretraining}
In this study, we hypothesize that graph information should be hierarchically structured, involving both nodes and motifs, and motifs serve as important functional subgraph patterns (such as functional groups in chemical molecules).
Given a graph ${G}$  with $\mathbf{X}$ and  $\mathbf{A}$, we aim to learn both an node level encoder $f_{\mathcal{N}}(\cdot)$ that produces node representations 
$\mathbf{H}^{\mathcal{N}}=f_{\mathcal{N}}(\mathbf{X, A})\in \mathbb{R}^{N_{\mathcal{N}}\times d_{\mathcal{N}}}$ 
and an subgraph level encoder $f_{\mathcal{M}}(\cdot)$ that generates motif representations
$\mathbf{H}^{\mathcal{M}}=f_{\mathcal{M}}(\mathbf{X, A})\in \mathbb{R}^{N_{\mathcal{M}}\times d_{\mathcal{M}}}$. 
Specifically, our goal is to autonomously discover motifs and then interactively learn node and motif representations, amalgamating both for diverse downstream tasks.

\subsection{The Design of DGPM}

\subsubsection{Dual-Level Pretrain}
\subsubsection{Node Feature Reconstruction Task}
For the node-level learning component, an encoder is designed primarily to capture the local node information. To this end, we employ a graph auto-encoder inspired by \cite{hou2022graphmae}, which prioritizes node feature reconstruction. The encoder transforms masked input data into a representation, and the decoder reverses this process to reconstruct the input, guided by the node feature reconstruction criterion.

Given $f_{\mathcal{N}}$ and $f'_{\mathcal{N}}$ as the encoder and decoder for node feature reconstruction, $\mathbf{H_{\mathcal{N}}}$ denoting the node representations encoded by $f_{\mathcal{N}}$, the goal of node level learning task is to reconstruct the input as 
\begin{align}
\mathbf{\widetilde{H}^{\mathcal{N}}} = f_\mathcal{N}(\mathbf{A,\widetilde{X}})\\
\mathbf{Z} = f'_\mathcal{N}(\mathbf{A,\widetilde{H}^{\mathcal{N}}})  
\end{align}
where $\mathbf{\widetilde{X}}$ and $\mathbf{Z}$ denote the masked and the reconstructed node features respectively.
Thus the reconstruction loss is as
\begin{align}
    \mathcal{L}^{\mathcal{N}}_{rec} = \frac{1}{|\widetilde{{V}}|} \sum_{v_i\in \widetilde{{V}}}(1-\frac{x^{\top}_i z_i} {||x_i||\cdot ||z_i||}).
\end{align}
\subsubsection{Subgraph Level Motif Auto-Discovery Task}
To effectively learn the subgraph structure information without human intervention, we propose a module for autonomous motif discovery, eliminating the need for domain-specific knowledge in motif predefinition. Taking inspiration from \cite{oliver2022approximate}, we customize EdgePool layers to merge nodes into subgraphs and enforce alignment between the cosine similarity of the merged subgraphs and the similarity generated by the graph kernel.

\textbf{EdgePool Layers}.
For an input graph, EdgePool layers selectively collapse pairs of nodes connected by an edge into a single node to obtain a coarsened graph. To automatically discover essential structures (motif), we train an aggregator and edge score function during this process. The aggregator combines connected node features, with edge features (if exist). The edge score function generates a score per edge to decide node-merge suitability.
Given $e_{u,v} $ as the edge to be merged, $f_{\mathcal{M}}(\cdot)$ as the aggregator, $s$ as the edge score, the aggregated representations and edge score are as:
\begin{align}
    h^{\mathcal{M}}_{u,v} & = f_{\mathcal{M}}(u,v,e_{u,v})  \nonumber \\
    & = [x_u\mathbf{W_{n}},x_v\mathbf{W_{n}}, x_{e_{u,v}}\mathbf{W_{e}}]\mathbf{W}
\end{align}
\begin{align}
    s_{u,v}= \sigma(h^{\mathcal{M}}_{u,v})    
\end{align}
where $\mathbf{W_n, W_e,}$ as transformation weight matrix and $\mathbf{W}$ are aggregation parameters, $x$ are corresponding node features and edge features, and $\sigma(\cdot)$ is the score function, a single-layer MLP is employed here.
Utilizing the generated edge scores, edges are sorted in reverse order of scores and sequentially compared with a uniformly distributed probability $p$. Edges with scores exceeding $p$ lead to the merging of the edge and connected nodes into a new node represented $h^{\mathcal{M}}_{u,v}$. As EdgePool layers stack, nodes in deep layers represent subgraphs derived from the input graph with informative structural features, thus completing the auto-discovery of the motif. Furthermore, as all pooling is edge-based, we can ensure the connectivity of the discovered motifs.

\textbf{Graph Similarity Loss}.
With EdgePool layers, the input graph is pooled into a coarsened graph, whose nodes denote motifs from the original graph. Hence, encoded node representations serve as corresponding motif representations. From a motif (subgraph) perspective, we adopt graph similarity as the training objective.
We employ the Wasserstein Weisfeiler Lehman (WWL) graph kernel\cite{togninalli2019wasserstein} to measure motif similarity as ground truth, guiding EdgePool layer training. Graph kernels excel in addressing graph complexity and exhibit good predictive capabilities across diverse graph tasks \cite{shervashidze2011weisfeiler,yanardag2015deep}. WWL graph kernel jointly models structural similarity and node feature agreement on graphs, effectively supervising graph topology properties.
For node pairs within the coarsened graph, we compute cosine similarity with encoded representations, aiming to align it with WWL kernel-generated similarity.
As shown in Fig~\ref{fig:model_framework}, given motif pair $(A,B)$ from motif graph $\mathcal{G}$,
\begin{align}
    \mathcal{L}^{\mathcal{M}}_{sim} = \sum_{\mathcal{G}}||\Omega(h^{\mathcal{M}}_A, h^{\mathcal{M}}_B)-\textbf{WWL}(\text{S}\{A\},\text{S}\{B\})||^2_2.
\end{align}
where $\Omega$ refers to scaled cosine similarity and $\text{S}\{\cdot\}$ as the corresponding motif of the node. 

\subsubsection{Cross-Level Matching Task}
Following training in node reconstruction and motif discovery, we obtain a node-level encoder and a subgraph-level encoder for 
corresponding representations. To exploit the inherent inter-relationship between nodes and motifs, we establish a node-motif matching task connecting node-level and subgraph-level training.
As shown in Figure~\ref{fig:model_framework}, the learned motif comprises distinct nodes, serving as the learning objective for the node-motif relationship. With permutation and concatenation, we iteratively combine node and motif representations and train a discriminator to predict whether an affiliation exists. Given permutation $\mathcal{P}$, corresponding matching labels $\mathbf{y}=\{y_{1,1},...,y_{1,N^{\mathcal{M}}},y_{2,1},...,y_{N^{\mathcal{N}},N^{\mathcal{M}}}\}$, and discriminator $g$, the matching loss is as
\begin{align}
    \mathcal{L}_{cross} & =\frac{1}{|\mathcal{P}|}\sum_{(i,j)\in \mathcal{P}}-\biggl(y_{i,j}\cdot 
    \log(g(h^\mathcal{N}_i,h^{\mathcal{M}}_j)) \nonumber \\
    &\quad +(1-y_{i,j})\cdot \log(1-g(h^\mathcal{N}_i,h^{\mathcal{M}}_j))\biggr)
\end{align}
Different from prior joint learning or step-by-step learning approaches with simple loss stacking, the proposed cross-level matching learning task establishes an \textit{interactive} connection between node-level encoder and subgraph-level encoder learning. In the training process, the node-level and subgraph-level representations can be iteratively enhanced based on the back-propagation of matching loss.

The overall training time complexity of DGPM is $\mathcal{O}(|V|\log(|V|))$, $|V|$ for the number of nodes, and details are shown in Appendix.D.

%% file: contents/5experiments.tex
\section{Experiments}

\input{contents/tabs/table1}
\input{contents/tabs/table2}

\subsection{Performance Validation}
To validate the performance of DGPM, we conducted experiments in two typical scenarios for downstream task applications: \textbf{\textit{Unsupervised Representation Learning}} (direct utilization of trained representations for graph classification) and \textbf{\textit{Transfer Learning}} (applying a pretrain-finetune approach for molecular property prediction). We followed the experimental setup employed in previous research work, such as data splits and evaluation metrics. Specifically, for unsupervised representation learning task, we adopted the experimental setup from \cite{zhang2021canonical,hou2022graphmae}; for transfer learning task, we followed the setup established in~\cite{hu2019strategies,you2020graph,you2021graph}.

\subsubsection{Datasets}
To validate \textit{unsupervised representation learning}, we conducted experiments on 7 graph classification benchmarks~\cite{hou2022graphmae}: MUTAG, IMDB-B, IMDB-M, PROTEINS, COLLAB, REDDIT-B, and NCI1. These datasets come from different fields and each of them is a collection of graphs where each graph is associated with a label. Specifically, node types serve as input features for MUTAG, PROTEINS, and NCI1 datasets, while node degrees are utilized in IMDB-B, IMDB-M, REDDIT-B, and COLLAB datasets.

To validate \textit{transfer learning}, we conduct molecular property prediction experiments under a pretrain-finetune setting. 250k unlabeled molecules sampled from the ZINC15\cite{sterling2015zinc} are used for pretraining and 8 molecular benchmark datasets~\cite{wu2018moleculenet} are used for finetuning and testing: BBBP, Tox21, ToxCast, SIDER, ClinTox, MUV, HIV, and BACE. The downstream datasets are partitioned using scaffold-split to emulate real-world scenarios. The input node features include the atom number and chirality tag, while the edge features encompass the bond type and direction.
Detailed characteristics of each dataset are available in Appendix A.
\subsubsection{Baselines}
For \textit{unsupervised representation learning} setting, we compared DGPM with two groups of strong unsupervised learning models. Graph kernel methods: Weisfeiler-Lehman sub-tree kernel (WL)~\cite{shervashidze2011weisfeiler} and Deep Graph Kernel (DGK)~\cite{yanardag2015deep}. SOTA self-supervised methods: graph2vec~\cite{narayanan2017graph2vec}, Infograph~\cite{sun2019infograph}, GraphCL~\cite{you2020graph}, JOAO~\cite{you2021graph}, GCC~\cite{qiu2020gcc}, MVGRL~\cite{hassani2020contrastive}, InfoGCL~\cite{xu2021infogcl}, and GraphMAE~\cite{hou2022graphmae}. 
Additionally, we have included two supervised methods as references to evaluate graph classification performance: GIN \cite{xu2018powerful} and DiffPool\cite{ying2018hierarchical}.

For \textit{transfer learning} setting, we compared DGPM with 7 self-supervised methods: ContextPred, AttrMasking~\cite{hu2019strategies}, Infomax~\cite{sun2019infograph}, GraphCL~\cite{you2020graph}, JOAO~\cite{you2021graph}, GraphLoG\cite{xu2021self} and GraphMAE~\cite{hou2022graphmae}. Furthermore, we included a no-pretrain model for comparison, in which DGPM is utilized for molecular property prediction without pretraining.

For each task, all baseline results were self-reported results in previous studies, under the same experimental setup.
\subsubsection{Implementation Details}
Generally, we adopt a 5-layer GIN as encoder and a single-layer GIN as decoder for node-level pretraining module. The hidden dimension is set to 128 for both node and motif representations. The framework is trained using the AdamW optimizer for 100 epochs, with all implementations carried out using the PyTorch Geometric package. The learned node representations and motif representations are first pooled by mean-pooling readout function and then concatenated as dual-level representations.

For \textit{unsupervised representation learning} setting,  the generated representations are fed into a downstream LIBSVM\cite{chang2011libsvm} classifier as graph features to predict the graph label. We report the mean 10-fold cross-validation accuracy with standard deviation after 5 runs. For \textit{transfer learning} setting, we employ a 2-layer EdgePool as the encoder within the motif discovery module. Additionally, an MLP layer is incorporated to adapt the combined representations for molecular property prediction. We conduct experiments in 10 repetitions and provide the mean and standard deviation of ROC-AUC scores (\%) as results.

More details on model configuration and the implementation environment can be found in Appendix B.
\subsubsection{Results}
For \textit{unsupervised representation learning}, the results are shown in Table~\ref{tab:table1}. DGPM outperforms all unsupervised baselines across all datasets, exhibiting an average improvement of 6.79\% and a maximum improvement of 28.12\%. DGPM also achieved comparable task performance with supervised learning on 3 datasets and even outperformed supervised learning on 4 benchmarks. The results show that dual-level pretraining is effective in learning informative representations and has the potential for unsupervised representation learning tasks.

For \textit{transfer learning}, Table~\ref{tab:table2} shows that our model's performance on downstream tasks outperforms SOTA methods on 6 out of 8 tasks (with maximum 25\% improvement), and achieves the best average performance (with average 4.2\% improvement). These results further indicate the robust transferability of DGPM.

In summary, DGPM achieves remarkable performance in both unsupervised representation learning and transfer learning across 15 benchmarks. The consistent outcomes in these two task settings demonstrate DGPM's effectiveness and generalisability for a wide range of applications in various domains.

\subsection{Ablation Study}
\input{contents/tabs/ablation}
\input{contents/tabs/ablation-motif}

\subsubsection{Effect of Motif Auto-Discovery Task}
As shown in Table~\ref{tab:ablation}, by comparing task performance before and after removing the motif discovery module, we have 
the following observations. (1) Generally, the inclusion of motif discovery module proves beneficial for pretraining tasks. Both representation learning-oriented graph classification and pretrain-finetune-oriented transfer learning benefit from motif discovery, resulting in accuracy improvements ranging from 1.8\% to 5.2\%. These improvements could potentially be attributed to the fact that the motif discovery module effectively learns valuable information about subgraph structures. (2) Motif discovery can provide more substantial improvements for larger scale graph datasets as it may extract more informative structural information. For instance, in unsupervised representation learning, DGPM with the motif discovery module exhibits greater enhancements for REDDIT-B than MUTAG (REDDIT-B, with an average of 429.7 nodes per graph, comprises larger graphs than MUTAG). Similarly, in transfer learning, motif discovery module brings a greater improvement in BBBP than Tox21 (The graph scale in BBBP is larger than Tox21).

\subsubsection{Effect of Motif Discovery Criterion}
By replacing WWL with edit distance, we examined the impact of graph similarity metric employed in motif discovery. As Table \ref{tab:ablation} shows, WWL offers advantages over edit distance. As WWL not only quantifies structural similarity but also incorporates node feature agreement in graph modeling, thereby effectively supervising graph topology properties.

\subsubsection{Effect of Cross-Level Matching Task}
As shown in Table \ref{tab:ablation}, we observe a significant drop in performance when the cross matching task is not included, amounting to an absolute drop of 0.4\% - 1.5\%. Interestingly, the datasets that benefit more from motif discovery also tend to benefit more from cross-level matching. These observations affirm that the cross-matching task indeed facilitates the learning of the inherent relationship between node-level information and subgraph-level structure, thereby contributing to a more comprehensive graph pretraining.


\subsection{Sensitive Analysis}
\begin{figure}
    \centering
    \includegraphics[width=8.5cm]{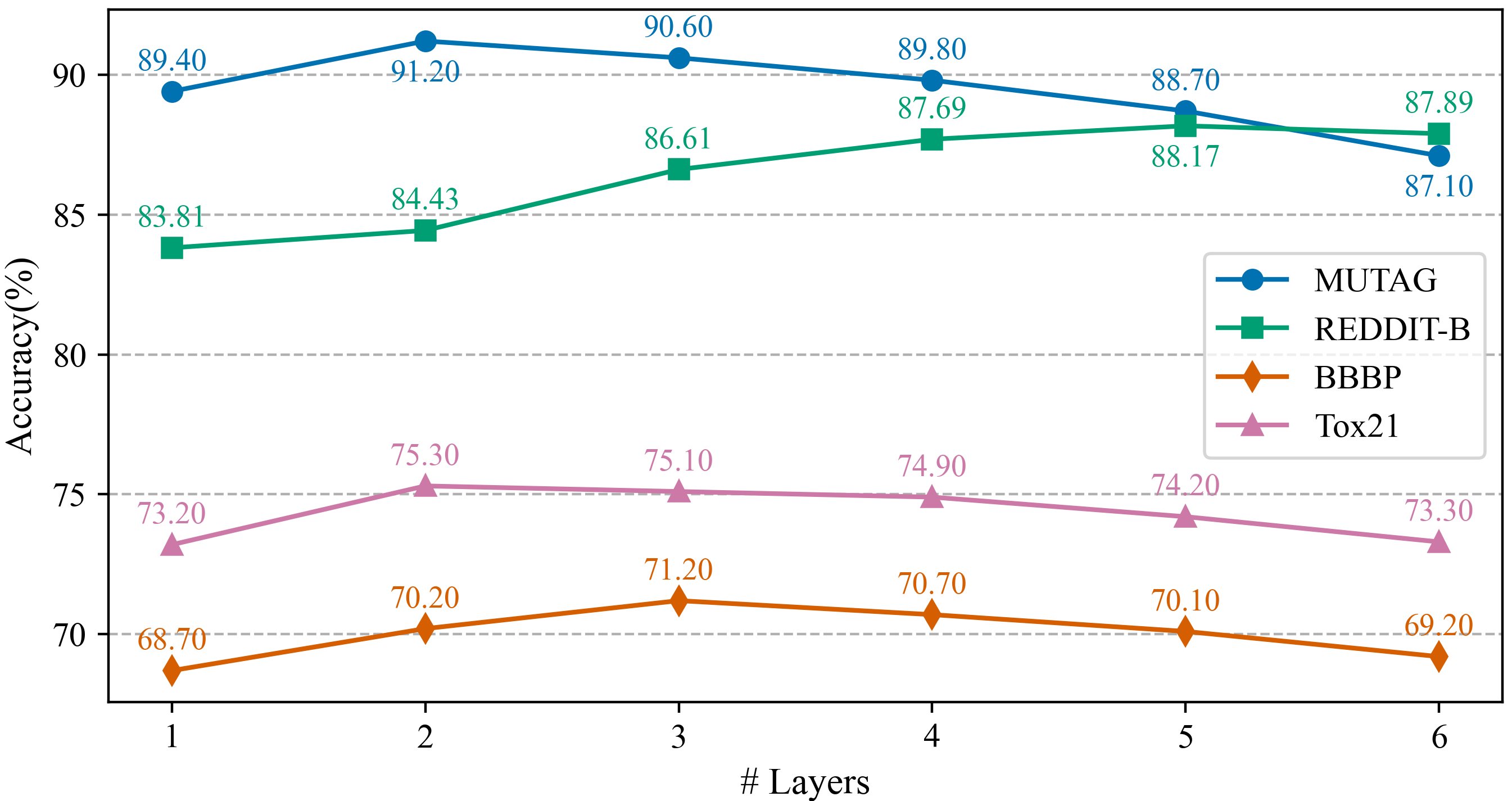}
    \caption{Sensitive analysis of the number of EdgePool layers in motif auto-discovery module.}
    \label{fig:sensitive}
\end{figure}

As nodes merge into subgraphs by the EdgePool layer in motif discovery module, the number of EdgePool layers $l$ decides the allowed maximum size of the learned subgraphs, i.e. motifs. Figure \ref{fig:sensitive} illustrates the impact of the number of EdgePool layers. Results for $l=1$ show significant underperformance across all four datasets. This indicates that, in most scenarios, the motif learning task with a single EdgePool layer is insufficient for effective motif discovery. For REDDIT-B, the model's performance steadily improves with increasing $l$, reaching 88.17\% at $l=5$. In the case of BBBP, the model achieves its peak performance at $l=2$, but its effectiveness decreases as $l$ grows larger. The optimal number of EdgePool layers varies across different datasets. This phenomenon is closely related to the motif properties in different graphs. Taking motifs in molecular graphs as an example, their sizes can range from 2, such as the hydroxyl group (-OH), to 6, like the benzene ring. With a limited number of EdgePool layers, the model struggles to learn more complex structures due to the constraint imposed by the maximum size of merged subgraphs. For larger graphs like those in REDDIT-B, intricate structures and large motifs are more prevalent, potentially leading the method to favor deeper networks.
Meanwhile, an excessive number of EdgePool layers may adversely affect the framework's performance. This is due to the power law relationship between the number of motifs in a graph and their average size. When the average motif size doubles, the number of motifs decreases by half. In the context of the cross-matching module, a lower number of motifs in a graph leads to less challenging discrimination tasks, which in turn hinders DGPM's learning performance.

\subsection{Analysis towards Motif Auto-Discovery}

\subsubsection{Comparison of Predefined Motif Methods}
To verify the superiority of the proposed motif auto-discovery module, we further simulate a practical scenario when the expert domain knowledge is insufficient and compare two graph pretrain methods using predefined motifs. As shown in Table~\ref{tab:ablation-motif}, for GROVER~\cite{rong2020self}, which uses a set of predefined motifs for pretrain, when we reduced the number of predefined motifs by half, its performance deteriorated across all datasets. Similarly, for MGSSL~\cite{zhang2021motif}, which applies manual filtering to extracted motifs by a molecule decomposition method (BRCIS), if we remove the manual filtering and directly use BRCIS-extracted motifs for pretrain, its performance also decreases on all datasets. 
These results show that the performance of graph pre-training methods, which depend on predefined motifs, can be easily influenced by the quality of these motifs.
In contrast, DGPM's autonomous motif discovery module eliminates the need for manual intervention, enhancing its stability and generalizability.
\subsubsection{Case Study}
\begin{figure}
    \centering
    \includegraphics[width=7.2cm]{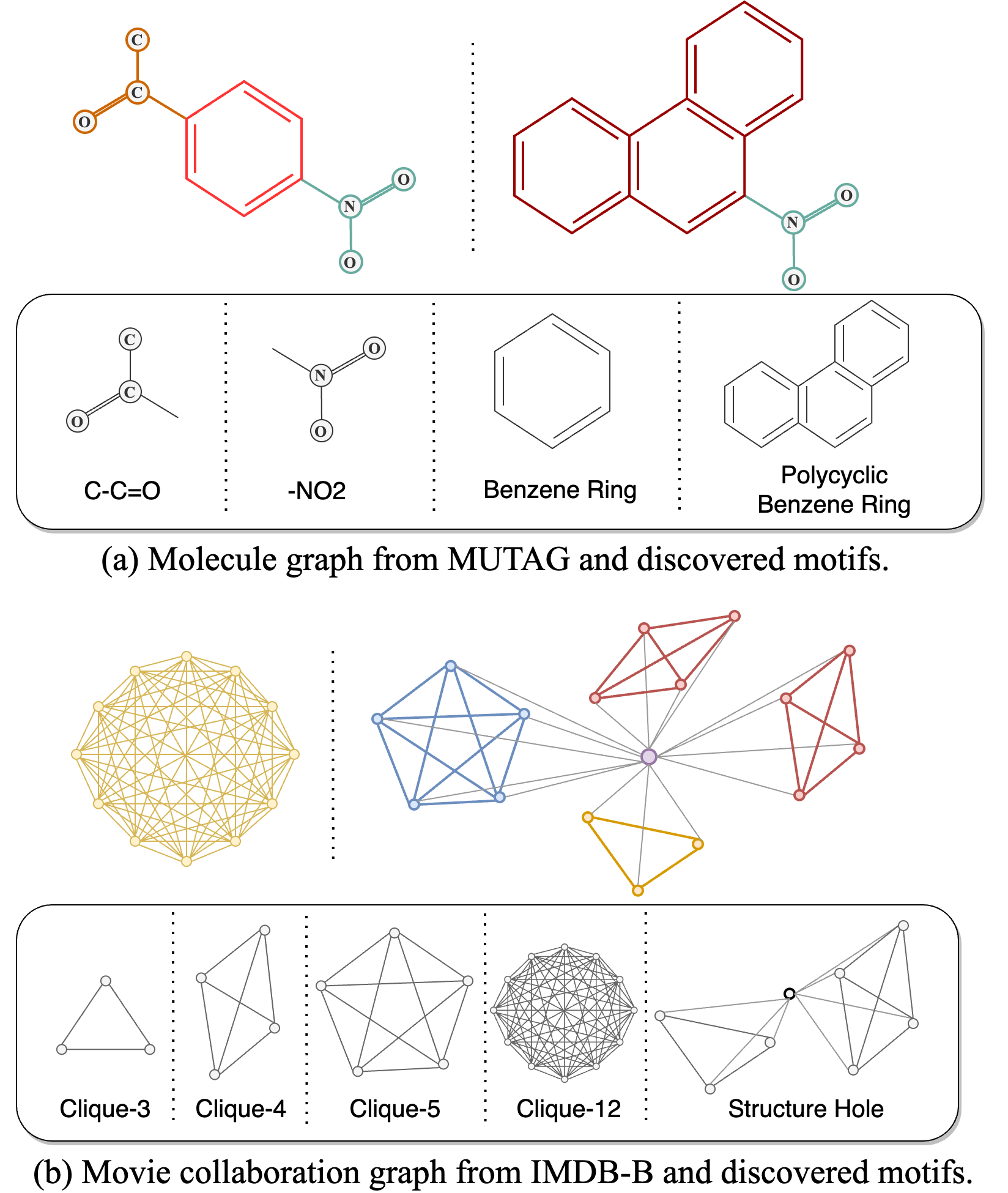}
    \caption{Auto-discovered motifs by DGPM from dataset MUTAG and IMDB-BINARY.}
    \label{fig:learned_motifs}
\end{figure}
For an input graph, motifs and their representations are learned simultaneously by motif auto-discovery in DGPM. This unique feature allows for direct visualization of the underlying motif representations, enhancing the interpretability of the learned representations. To illustrate this, we check discovered graph motifs from diverse domains, encompassing both natural science and social sciences, as depicted in Figure~\ref{fig:learned_motifs}. Additional auto-discovered motifs from various datasets are provided in Appendix C.

In MUTAG, the graphs represent nitroaromatic compounds, which are organic molecules containing at least one nitro group (-NO2) attached to a benzene ring. Figure~\ref{fig:learned_motifs} (a) illustrates the auto-discovered motifs of nitroaromatic compounds, including -NO2 and benzene rings, achieved by DGPM without incorporating chemical functional group information. Additionally, DGPM also identifies polycyclic aromatic structures that imply high mutagenic potency\cite{debnath1991structure}.

IMDB-B graphs represent movie collaboration networks, with nodes representing actors/actresses and edges denoting co-appearances in movies. IMDB-B can provide an example of social network analysis. Figure~\ref{fig:learned_motifs} (b) displays original collaboration networks along with motifs autonomously discovered by DGPM. These motifs vary in size but share similar clique structures. Cliques, which are complete subgraphs where every member is directly connected to every other, indicate strong connections within a group. Furthermore, Figure~\ref{fig:learned_motifs} (b) identifies a structure hole—an interconnected node that links multiple cliques into a larger collaboration network. Structure holes play a crucial role in social networks by bridging disconnected groups and facilitating information flow between them, fostering collaborations and new ideas~\cite{burt2001structural}.

Additionally, a comparison between auto-discovered motifs and those decomposed by rules defined in \cite{zhang2021motif} was conducted for the BBBP dataset. The results revealed that DPGM successfully discovered 64.28\% of the motifs, highlighting its proficiency in motif discovery for chemical graphs. Further elaboration on this comparison is provided in Appendix C.

%% file: contents/tabs/table1.tex

\begin{table*}[t]
\centering
\fontsize{9}{11}\selectfont
\begin{tabular}{@{}ccccccccc@{}}
\toprule
       Category                  & Method   & IMDB-B              & IMDB-M              & PROTEINS            & COLLAB            & MUTAG               & REDDIT-B              & NCI1                \\ \midrule
\multirow{2}{*}{Supervised}      & GIN       & 75.1±5.1            & 52.3±2.8            & 76.2±2.8            & 80.2±1.9          & 89.4±5.6            & 92.4±2.5            & 82.7±1.7            \\
                                 & DiffPool  & 72.6±3.9            & -                   & 75.1±3.5            & 78.9±2.3          & 85.0±10.3           & 92.1±2.6            & -                   \\ \midrule
\multirow{2}{*}{\parbox{2cm}{\centering Unsupervised\\ Graph Kernel}}    & WL        & 72.30±3.44          & 46.95±0.46          & 72.92±0.56          & 78.9±1.9          & 80.72±3.00          & 68.82±0.41          & 80.31±0.46          \\
                                 & DGK       & 66.96±0.56          & 44.55±0.52          & 73.30±0.82          & 73.1±0.3          & 87.44±2.72          & 78.04±0.39          & 80.31±0.46          \\ \cmidrule(l){2-9} 
\multirow{9}{*}{\parbox{2cm}{\centering Self-Supervised}} & graph2vec & 71.10±0.54          & 50.44±0.87          & 73.30±2.05          & -                 & 83.15±9.25          & 75.58±1.03          & 73.22±1.81          \\
                                 & Infograph & 73.03±0.87          & 49.69±0.53          & 74.44±0.31          & 70.65±1.13        & 89.01±1.13          & 82.50±1.42          & 76.20±1.06          \\
                                 & GraphCL   & 71.14±0.44          & 48.58±0.67          & 74.39±0.45          & 71.36±1.15        & 86.80±1.34          & 87.53±0.84          & 77.87±0.41          \\
                                 & JOAO      & 70.21±3.08          & 49.20±0.77          & 74.55±0.41          & 69.50±0.36        & 87.35±1.02          & 85.29±1.35          & 78.07±0.47          \\
                                 & GCC       & 72.0                & 49.4                & 74.48±3.12          & 78.9              & -                   & 89.8                & 66.33±2.65          \\
                                 & MVGRL     & 74.20±0.70          & 51.20±0.50          & 71.50±0.30          & 76.01±1.20        & 89.70±1.10          & 84.50±0.60          & -                   \\
                                 & InfoGCL   & 75.10±0.90          & 51.40±0.80          & -                   & 80.00±1.30        & \underline{91.20±1.30}    & -             & 80.20±0.60           \\
                                 & GraphMAE  & \underline{75.52±0.66}    & \underline{51.63±0.52}    & \underline{75.30±0.39}    & \underline{80.32±0.46}  & 88.19±1.26          & \underline{88.01±0.19}    & \underline{80.40±0.30}    \\ \cmidrule(l){2-9} 
                                 & DGPM     & \textbf{75.77±0.53} & \textbf{52.12±0.47} & \textbf{75.72±0.43} & \textbf{80.44±0.54} & \textbf{91.20±0.87} & \textbf{88.17±0.31} & \textbf{80.87±0.28} \\ \bottomrule
\end{tabular}
\caption{Experiment results in \emph{unsupervised representation learning} for graph classification. We report accuracy (\%) for all datasets. The reported results of baselines are from previous papers if available.}
\label{tab:table1}
\end{table*}

%% file: contents/tabs/table2.tex
\begin{table*}[t]
\fontsize{9}{11}\selectfont
\centering
\begin{tabular}{@{}cccccccccc@{}}
\toprule
            & BBBP              & Tox21             & ToxCast        & SIDER          & ClinTox        & MUV            & HIV               & BACE           & Avg.  \\ \midrule
No-pretrain & 66.5±1.4          & 74.7±0.4          & 63.3±1.5       & 56.4±0.8       & 58.6±2.1       & 71.9±1.4       & 75.4±0.8          & 72.3±1.6       & 67.39 \\ \midrule
ContextPred & 63.4±2.7          & 74.7±0.7          & 62.9±0.6       & 59.0±0.6       & 64.7±3.8       & 74.2±0.9       & 75.4±1.1          & 78.9±2.1       & 69.15 \\
AttrMasking & 63.5±2.7          & \textbf{75.4±0.6} & 63.2±0.5       & 59.1±0.7       & 70.7±2.1       & 73.1±2.2       & 75.2±1.0          & 78.7±1.7       & 69.86 \\
Infomax     & 66.7±0.8          & 74.3±0.5          & 61.5±0.5       & 57.3±0.9       & 68.1±3.1       & 73.4±2.6       & 74.2±1.2          & 74.7±1.6       & 68.78 \\
GrpahCL     & 67.8±0.8          & 72.7±0.9          & 60.4±0.6       & 58.6±0.8       & 74.2±2.1       & 68.9±3.2       & \underline{77.1±0.9}    & 74.1±0.7       & 69.23 \\
JOAO        & 69.1±1.2          & 73.8±0.5          & 61.1±0.4       & 58.7±0.8       & 80.1±1.8       & 70.2±1.1       & 75.1±0.8          & 75.6±0.8       & 70.46 \\
GraphLoG    & \underline{70.4±0.8}    & 74.9±0.5          & 61.8±0.7       & \underline{59.5±1.1} & 75.7±2.6       & \underline{74.8±1.1} & 74.9±0.9          & 81.0±1.1       & 71.63 \\
GraphMAE    & 70.1±0.6          & 74.4±0.5          & \underline{63.9±0.4} & 59.0±0.7       & \underline{80.8±1.2} & 74.5±2.3       & 77.0±0.4          & \textbf{81.4±0.9}      & 72.64 \\ \midrule
DGPM       & \textbf{71.2±0.5} & \underline{75.3±0.4}    & \textbf{64.0±0.7}       & \textbf{60.3±0.8}       & \textbf{80.9±1.3}       & \textbf{75.3±1.6}       & \textbf{77.3±0.6} & \underline{81.1±0.7} & \textbf{73.17} \\ \bottomrule
\end{tabular}
\caption{Experiment results in \emph{transfer learning} on molecular property prediction benchmarks. The model is first pre-trained on ZINC15 and then finetuned on the following datasets. We report ROC-AUC scores (\%).}
\label{tab:table2}
\end{table*}


%% file: contents/tabs/ablation.tex
\begin{table}
\fontsize{9}{11}\selectfont
\centering
\setlength{\tabcolsep}{2pt}
\begin{tabular}{@{}ccccc@{}}
\toprule
Task               & \multicolumn{2}{c}{Representation Learning} & \multicolumn{2}{c}{Transfer Learning} \\ \cmidrule(l){2-5} 
Dataset            & MUTAG             & REDDIT-B             & BBBP              & Tox21             \\ \midrule
DGPM              & 91.20              & 88.17                & 71.2              & 75.3              \\
w/o cross matching & 90.80              & 87.42                & 69.7             & 74.4              \\
w/ edit distance   & 89.67             & 85.03                & 69.0             & 74.3              \\
w/o motif discovery & 88.19             & 84.29                & 67.4             & 72.7              \\ \bottomrule
\end{tabular}
\caption{Ablation studies of the cross matching, motif discovery and measurement of motif similarity.}
\label{tab:ablation}
\end{table}

%% file: contents/tabs/ablation-motif.tex
\begin{table*}[t]
\fontsize{9}{11}\selectfont
\centering
\begin{tabular}{@{}cccccccccl@{}}
\toprule
                     & BBBP  & Tox21 & ToxCast & SIDER & ClinTox & MUV   & HIV   & BACE  & Avg.   \\ \midrule
GROVER               & 68.03 & 76.31 & 63.39    & 60.66 & 76.92   & 75.78 & 77.81 & 79.51 & 72.30 \\ 
GROVER(half \#motifs) & 67.11↓ & 74.67↓ & 62.11↓   & 60.04↓ & 74.25↓   & 74.87↓ & 75.35↓ & 76.87↓ & 70.66↓ \\ \midrule
MGSSL                & 69.68  & 76.36 & 64.12   & 61.81 & 79.98   & 78.68 & 78.72 & 79.12 & 73.56 \\
MGSSL(BRCIS)         & 67.52↓ & 73.62↓ & 62.34↓   & 59.24↓ & 77.1↓    & 77.63↓ & 76.34↓ & 75.63↓ & 71.18↓ \\ \midrule
DGPM(auto-discovered)               & 71.15 & 75.28 & 64.02   & 60.3  & 80.91   & 75.28 & 77.26 & 81.13 & 73.17 \\ \bottomrule
\end{tabular}
\caption{Comparison of DGPM and predefined-motif graph self-supervised methods. These models are first pre-trained on ZINC15 and then finetuned on the following datasets. We report ROC-AUC scores (\%). For GROVER, we compare with the model pretrained with half number of motifs in designed motif prediction task. For MGSSL, we compare with the model trained with motifs generated by the molecule decomposition method BRCIS and without further processing.}
\label{tab:ablation-motif}
\end{table*}

%% file: contents/2relatedWork.tex
\section{Related Works}

\subsection{Self-Supervised Graph Pretraining}
Self-supervised pretraining allows the model to acquire universal knowledge from large-scale unlabeled datasets and offers a superior starting point for downstream tasks \cite{hao2019visualizing,zoph2020rethinking}. Depending on pretext task design, self-supervised pretraining methods on graphs can be categorized as contrastive or predictive learning.
\textbf{Constrastive learning}. Given training graphs, contrastive learning aims to learn graph encoders such that representations of similar graph instances exhibit concordance while representations of dissimilar instances manifest disagreement.
DGI \cite{velivckovic2018deep} and InfoGraph \cite{sun2019infograph} adopt the local-global mutual information maximization to learn node-level and graph-level representations. MVGRL \cite{hassani2020contrastive} leverages graph diffusion to generate an additional view and contrasts node-graph representations of distinct views. GCC \cite{qiu2020gcc} utilizes discrimination of subgraph-level instances generated by ego-graph for the pre-training. GRACE \cite{zhu2020deep}, GraphCL \cite{you2020graph}, GCA \cite{zhu2021graph}, D-SLA \cite{kim2022graph} learn the node or graph representation by maximizing the agreement between different augmentations. GGD \cite{zheng2022rethinking} analyzes the defect of existing contrastive learning methods and introduces a group discrimination paradigm. 
\textbf{Predictive learning}. Compared with contrastive learning, predictive learning methods train the graph encoder $f$ together with a prediction head $f'$, guided by self-generated informative labels. Graph auto-encoders (GAEs) adhere to the essence of auto-encoders \cite{hinton1993autoencoders}, by reconstructuring input graph to learn node representations. Most GAEs include reconstructing structural information \cite{pan2018adversarially,wang2017mgae,park2019symmetric,tang2022graph} by link prediction. GraphMAE \cite{hou2022graphmae} focuses on reconstructing node features with masked graphs. Its successor, GraphMAE2 \cite{hou2023graphmae2}, further refines the framework via a multi-view random re-mask mechanism.
In addition to graph auto-encoders, inspired by the success of autoregressive models in natural language processing, GPT-GNN \cite{hu2020gpt} designs an attributed graph generation task, including attribute and edge generation, for pretraining GNN models.


Although numerous methods try to leverage graph structural information in self-supervised pretraining, the utilization of subgraph-level patterns, e.g., motif, remains relatively unexplored, both in terms of graph view augmentation and pretext task formation.


\subsection{Motif-Enhanced Graph Pretraining}
Graph motifs are patterns of interconnections occurring in complex networks at numbers that are significantly higher than those in randomized networks \cite{milo2002network}. Serving as fundamental units of graphs, motifs reveal interconnections of nodes and offer insight into the entire graph's characteristics. Hence, motifs hold the potential to enhance the performance of graph pre-training models. Several studies have initiated the design of motif-based self-supervised tasks that incorporate motif semantics to acquire more informative graph representations. GROVER \cite{rong2020self} utilizes motifs (functional groups) in the molecule, which is extracted by RDKit, as property prediction labels for pretraining tasks. MMGSL \cite{zhang2021motif} introduces a motif generation task for molecule graphs, decomposing motifs using the BRICS algorithm along with human-designed rules.

However, motifs in existing works are usually predefined based on domain-specific knowledge, thus these models have a high dependency on human intervention and can only be applied to specific domains, constraining the methods' generalizability.

%% file: contents/6conclusion.tex
\section{Conclusion and Future Work}
In this study, to address the challenges encountered in graph pretraining, we propose DGPM, a dual-level graph self-supervised pretraining with motif discovery. DGPM introduces a motif auto-discovery task to effectively learn subgraph-level topological information. A cross-matching learning module is proposed for better dual-level feature fusion. Comprehensive experiments conducted across diverse graph learning benchmarks demonstrate the effectiveness and generalizability of DGPM\footnote{Supplementary materials on the model and experiments can be found at https://github.com/RocccYan/DGPM.}.
Future directions could involve exploring the extraction of heterogeneous graph's motifs and the automated learning of hyperparameters.

%% file: contents/7appendix.tex
\onecolumn
\appendix
\section{Appendix}

\subsection{A. Datasets}
\subsubsection{Datasets for Unsupervised Graph Classification}
The datasets for unsupervised graph classification are selected from different domains. Introductions of these datasets are as below and their statistics are shown in Table \ref{tab:data-unsupervised}:
\begin{itemize}
    \item IMDB-BINARY \cite{velivckovic2018deep}: IMDB-BINARY(IMDB-B) is a movie collaboration dataset that consists of the ego-networks of 1,000 actors/actresses who played roles in movies in IMDB. In each graph, nodes represent actors/actress, and there is an edge between them if they appear in the same movie. These graphs are derived from the Action and Romance genres.
    \item IMDB-MULTI \cite{velivckovic2018deep}: IMDB-MULTI is a relational dataset that consists of a network of 1000 actors or actresses who played roles in movies in IMDB. A node represents an actor or actress, and an edge connects two nodes when they appear in the same movie. In IMDB-MULTI, the edges are collected from three different genres: Comedy, Romance and Sci-Fi.
    \item PROTEINS \cite{borgwardt2005protein}: PROTEINS is a dataset of proteins that are classified as enzymes or non-enzymes. Nodes represent the amino acids and two nodes are connected by an edge if they are less than 6 Angstroms apart.
    \item COLLAB \cite{velivckovic2018deep}: COLLAB is a scientific collaboration dataset. A graph corresponds to a researcher’s ego network, i.e., the researcher and its collaborators are nodes and an edge indicates collaboration between two researchers. A researcher’s ego network has three possible labels, i.e., High Energy Physics, Condensed Matter Physics, and Astro Physics, which are the fields that the researcher belongs to. The dataset has 5,000 graphs and each graph has label 0, 1, or 2.
    \item MUTAG \cite{debnath1991structure}: MUTAG is a collection of nitroaromatic compounds and the goal is to predict their mutagenicity on Salmonella typhimurium. Input graphs are used to represent chemical compounds, where vertices stand for atoms and are labeled by the atom type (represented by one-hot encoding), while edges between vertices represent bonds between the corresponding atoms. It includes 188 samples of chemical compounds with 7 discrete node labels.
    \item REDDIT-B \cite{velivckovic2018deep}: REDDIT-BINARY (REDDIT-B) consists of graphs corresponding to online discussions on Reddit. In each graph, nodes represent users, and there is an edge between them if at least one of them respond to the other’s comment. There are four popular subreddits, namely, IAmA, AskReddit, TrollXChromosomes, and atheism. IAmA and AskReddit are two question/answer based subreddits, and TrollXChromosomes and atheism are two discussion-based subreddits. A graph is labeled according to whether it belongs to a question/answer-based community or a discussion-based community.
    \item NCI1 \cite{wale2008comparison}: The NCI1 dataset comes from the cheminformatics domain, where each input graph is used as representation of a chemical compound: each vertex stands for an atom of the molecule, and edges between vertices represent bonds between atoms. This dataset is relative to anti-cancer screens where the chemicals are assessed as positive or negative for cell lung cancer. Each vertex has an input label representing the corresponding atom type, encoded by a one-hot-encoding scheme into a vector of 0/1 elements.
\end{itemize}

\subsubsection{Datasets for Transfer Learning}
The datasets for transfer learning include ZINC \cite{irwin2012zinc} for pretraining and eight downstream datasets from MoleculeNet \cite{wu2018moleculenet}.
ZINC is a free database of commercially-available compounds for virtual screening. ZINC contains over 230 million purchasable compounds in ready-to-dock, 3D formats. ZINC also contains over 750 million purchasable compounds that can be searched for analogs. For simplicity, we use a minimal set of node and bond features that unambiguously describe the two-dimensional structure of molecules. We use RDKit \cite{landrum2006rdkit} to obtain these features.
\begin{itemize}
    \item Node features:
    \begin{itemize}
        \item Atom number: [1, 118]
        \item Chirality tag: \{unspecified, tetrahedral cw, tetrahedral ccw, other\}
    \end{itemize}
    \item Edge features:
    \begin{itemize}
        \item Bond type: {single, double, triple, aromatic}
        \item Bond direction: {–, endupright, enddownright}
    \end{itemize}
\end{itemize}

Eight binary graph classification datasets are used to evaluate model performance and here are the introductions:
\begin{itemize}
    \item BBBP \cite{martins2012bayesian}. Blood-brain barrier penetration (membrane permeability).
    \item Tox21 \cite{mayr2016deeptox}. Toxicity data on 12 biological targets, including nuclear receptors and stress response pathways.
    \item ToxCast \cite{richard2016toxcast}. Toxicology measurements based on over 600 in vitro high-throughput screenings.
    \item SIDER \cite{kuhn2016sider}. Database of marketed drugs and adverse drug reactions (ADR), grouped into 27 system organ classes.
    \item ClinTox \cite{novick2013sweetlead}. Qualitative data classifying drugs approved by the FDA and those that have failed clinical trials for toxicity reasons.
    \item MUV \cite{gardiner2011effectiveness}. Subset of PubChem BioAssay by applying a refined nearest neighbor analysis, designed for validation of virtual screening techniques.
    \item HIV \cite{riesen2008iam}. Experimentally measured abilities to inhibit HIV replication.
    \item BACE \cite{subramanian2016computational}. Qualitative binding results for a set of inhibitors of human $\beta$-secretase 1.
\end{itemize}

The statistics of datasets for transfer learning molecular property prediction are shown in Table \ref{tab:data-transfer}.

\input{contents/tabs/app-data_unsupervised}
\input{contents/tabs/app-data_transfer}
\input{contents/tabs/app-MC_unsupervised}
\input{contents/tabs/app-MC_transfer}
\subsection{B. Implementation}
\subsubsection{Running Environment}
All experiments are conducted on Linux servers equipped with an Intel(R) Xeon(R) Platinum 8163 CPU @ 2.50GHz, 88GB RAM, and NVIDIA V100 GPUs. Models of pretraining and downstream tasks are implemented in \emph{PyTorch} version 1.8.1, \emph{Pytorch Geometric} 2.2.0 with \emph{CUDA} version 10.1, \emph{scikit-learn } version 0.24.1 and \emph{Python} 3.7. For node feature reconstruction, we implement our model based on the code in \emph{https://github.com/THUDM/GraphMAE}. For molecular property prediction, we implement our model based on the code in \emph{https://github.com/snap-stanford/pretrain-gnns}. 

\subsubsection{Model Configuration}
For unsupervised representation learning, we search the optimal number of EdgePool layers from $\{2,3,4,5,6,7,8\}$ with the allowed maximum size of motifs are from $2^2 $ to $2^8$ for different datasets.
For evaluation, the parameter $c$ of SVM is searched in the sets $\{10^{-3}, 10^{-2}, 0.1, 1, 10\}$. The detailed hyper-parameters by datasets are shown in Table \ref{tab:MC-unsupervised}.
For transfer learning of molecule property prediction, we search the optimal number of EdgePool layers from $\{2,3,4,\}$.
Besides, we adopt a single layer MLP as discriminator to adapt with different number of prediction tasks among downstream datasets.
The detailed model configurations of transfer learning molecular property prediction are shown in Table \ref{tab:MC-tranfer}.

\subsection{C. Analysis of Auto-Discovered Motif from DGPM }
\subsubsection{Discovered Motifs from Datasets of Different Fields}
\begin{figure}
    \centering
    \includegraphics[width=8cm]{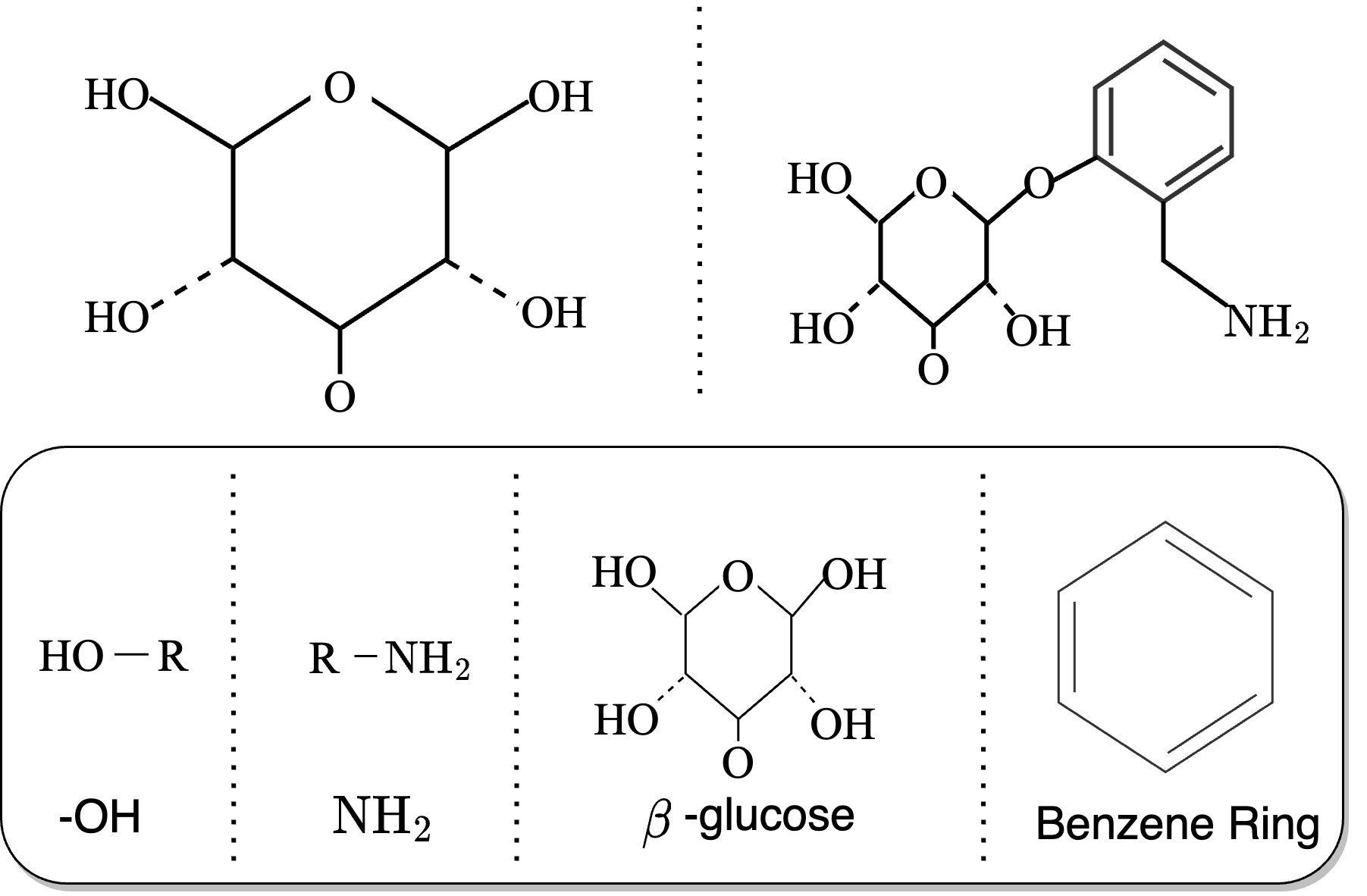}
    \caption{Molecule graph from BBBP and discovered motifs.}
    \label{fig:app-motif}
\end{figure}

\begin{figure}
    \centering
    \includegraphics[width=18cm]{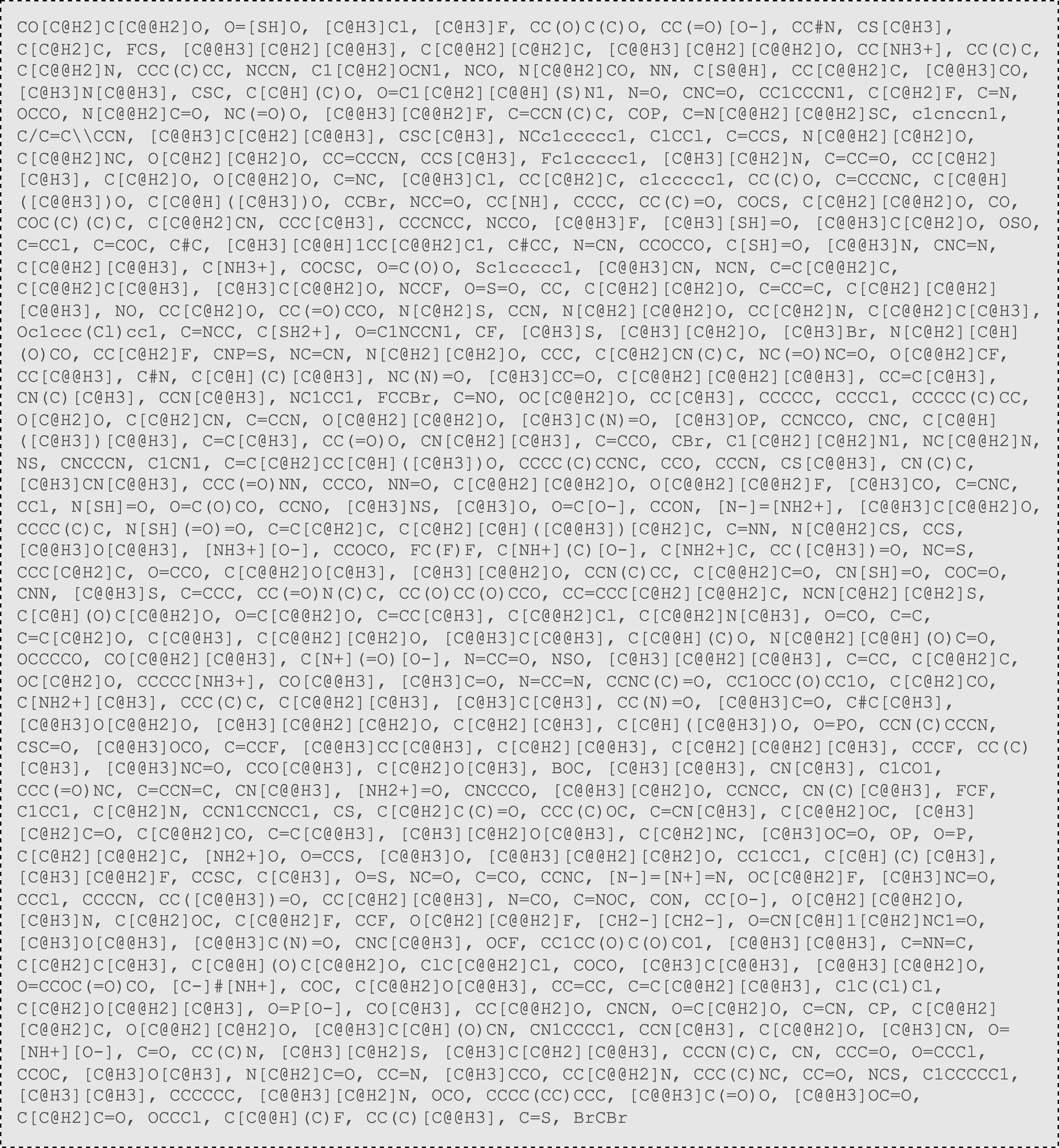}
    \caption{SMILES for discovered motifs from BBBP.}
    \label{fig:app-smiles}
\end{figure}

Additional to MUTAG and IMDB-BINARY, here we visualize cases of the learned motifs in Figure \ref{fig:app-motif} and also, show the SMILES for discovered motifs from BBBP in Figure \ref{fig:app-smiles}.

\subsubsection{Comparison Between Auto-Discovered Motifs with Defined Motifs}
To explicitly validate the ability of motif auto-discovery module, we design a comparison experiment to test the coverage for the motifs learned by this module on motifs decomposed by professional tools and human-defined rules. The comparison is based on SMILES of motifs generally utilized to represent chemical molecule structures. Taking dataset BBBP for an example, here are the detailed procedures:
\begin{enumerate}[label=\Roman*.]
    \item Train DGPM framework on ZINC as introduced in transfer learning setting.
    \item Apply the pretrained EdgePool layers in Motif Auto-discovery module to dataset BBBP to generate motifs for each molecule graph.
    \item Based on the marked motif labels of graphs in dataset BBBP, break each graph into a set of subgraphs, then generating a dataset of subgraphs can be called as motif dataset.
    \item With the support of \texttt{torch\_geometric.utils.smiles} module in package Pytorch Geometric, map each subgraph, i.e. motif, in motif dataset to corresponding SMILES and then get the set of SMILES from learned motifs.
    \item According to \cite{zhang2021motif}, generate the SMILES of motifs which are decomposed by professional tool BRCIS and human-defined rules, as the ground truth.
    \item Given the ground truth motifs and auto-discovered motifs represented by SMILES, we check the intersection of these two and take the number of motifs in the intersection divided by the number of ground truth motifs as the reported coverage of auto-discovered motifs.
\end{enumerate}

\subsection{D. Complexity Analysis}
In this section, we analyze the time complexity of DGPM by three components, i.e. node level feature reconstruction, subgraph level motif discovery, and cross matching between nodes and motifs, and report the typical running time for training and inference. 
For node feature reconstruction, the time complexity is linear
with the number of nodes $n$, that is $\mathcal{O}(n)$. For motif auto-discovery, as equation \(6\) in the main text shows, the similarity loss is measured by combination pairs of motifs, and the time complexity is $\mathcal{O}(m^2)$. Then for cross matching part, as equation \(7\) in the main text shows, the cross matching loss is computed by permutation of nodes and motifs, thus having a time complexity of $\mathcal{O}(nm)$.
Given motifs are merged by EdgePool layers, we have $m\sim \log(n)$, thus the time complexity of the DGPM framework is as $\mathcal{O}(n\log(n))$. The typical pretraining time on molecule graphs in size of 1,000 is around 2 hours and the inference time takes about 10 seconds. 

%% file: contents/tabs/app-data_unsupervised.tex
\begin{table}[]
\centering

\begin{tabular}{cccccccc}
\toprule
Statistics& IMDB-B & IMDB-M & PROTEINS & COLLAB & MUTAG & REDDIT-B & NCI1  \\ \midrule
\# graphs     & 1,000  & 1,500  & 1,113    & 5,000  & 188   & 2,000    & 4,110 \\
\# classes    & 2      & 3      & 2        & 3      & 2     & 2        & 2     \\
Avg. \# nodes & 19.8   & 13     & 39.1     & 74.5   & 17.9  & 429.7    & 29.8  \\ \bottomrule
\end{tabular}
\caption{Statistics for unsupervised graph classification datasets.}
\label{tab:data-unsupervised}
\end{table}

%% file: contents/tabs/app-data_transfer.tex
\begin{table}[]
\centering
\begin{tabular}{cccccccccc}
\toprule
Statistics               & ZINC    & BBBP  & Tox21 & ToxCast & SIDER & ClinTox & MUV    & HIV    & BACE  \\
\midrule
\# graphs             & 250,000 & 2,039 & 7,831 & 8,576   & 1,427 & 1,477   & 93,087 & 41,127 & 1,513 \\
\#   prediction tasks & -       & 1     & 12    & 617     & 27    & 2       & 17     & 1      & 1     \\
Avg. \# nodes         & 26.6    & 24.1  & 18.6  & 18.8    & 33.6  & 26.2    & 24.2   & 24.5   & 34.1  \\
\bottomrule
\end{tabular}
\caption{Statistics for pretrain-finetune molecular property prediction. \emph{ZINC} is for pretraining and others are for downstream prediction.}
\label{tab:data-transfer}
\end{table}

%% file: contents/tabs/app-MC_unsupervised.tex
\begin{table}[]
\centering

\begin{tabular}{cccccccc}
\toprule
Hyper-parameters   & IMDB-B & IMDB-M & PROTEINS & COLLAB & MUTAG & REDDIT-B & NCI1  \\
\midrule
masking rate       & 0.50   & 0.50   & 0.50     & 0.75   & 0.75  & 0.75     & 0.25  \\
\# EdgePool layers & 3      & 3      & 3        & 5      & 3     & 6        & 3     \\
hidden size        & 128    & 128    & 128      & 128    & 128   & 128      & 128   \\
max epoch          & 100    & 100    & 100      & 200    & 100   & 200      & 100   \\
batch size         & 256    & 256    & 256      & 128    & 256   & 128      & 256   \\
pooling            & mean   & mean   & mean     & mean   & mean  & mean     & mean  \\
learning rate      & 0.005  & 0.001  & 0.005    & 0.005  & 0.005 & 0.001    & 0.005 \\
weight decay       & 2e-04  & 2e-05  & 2e-04    & 1e-04  & 2e-04 & 2e-05    & 2e-04 \\
\bottomrule
\end{tabular}
\caption{Hyper-parameters of unsupervised graph classification experiments.}
\label{tab:MC-unsupervised}
\end{table}

%% file: contents/tabs/app-MC_transfer.tex
\begin{table}[]
\centering

\begin{tabular}{cccccccccc}
\toprule
Hyper-parameters   & ZINC  & BBBP  & Tox21 & ToxCast & SIDER & ClinTox & MUV   & HIV   & BACE  \\
\midrule
masking rate       & 0.50  & -     & -     & -       & -     & -       & -     & -     & -     \\
\# EdgePool layers & 2     & 2     & 2     & 2       & 2     & 2       & 2     & 2     & 2     \\
hidden size        & 128   & 128   & 128   & 128     & 128   & 128     & 128   & 128   & 128   \\
\# MLP layers      & -     & 1     & 1     & 1       & 1     & 1       & 1     & 1     & 1     \\
max epoch          & 100   & 10    & 10    & 10      & 10    & 10      & 10    & 10    & 10    \\
batch size         & 256   & 64    & 128   & 128     & 32    & 32      & 128   & 128   & 32    \\
pooling            & mean  & mean  & mean  & mean    & mean  & mean    & mean  & mean  & mean  \\
learning rate      & 0.005 & 0.001 & 0.001 & 0.001   & 0.001 & 0.001   & 0.001 & 0.001 & 0.001 \\
weight decay       & 2e-04 & 1e-04 & 1e-04 & 1e-04   & 1e-04 & 1e-04   & 1e-04 & 1e-04 & 1e-04 \\
\bottomrule
\end{tabular}
\caption{Hyper-parameters of transfer learning molecular property prediction experiments.}
\label{tab:MC-tranfer}
\end{table}